\documentclass[pdflatex,sn-mathphys-num]{sn-jnl}

\usepackage{graphicx}%
\usepackage{multirow}%
\usepackage{amsmath,amssymb,amsfonts}%
\usepackage{amsthm}%
\usepackage{mathrsfs}%
\usepackage[title]{appendix}%
\usepackage{xcolor}%
\usepackage{textcomp}%
\usepackage{manyfoot}%
\usepackage{booktabs}%
\usepackage{algorithm}%
\usepackage{algorithmicx}%
\usepackage{algpseudocode}%
\usepackage{listings}%

\theoremstyle{thmstyleone}%
\theoremstyle{thmstyletwo}%

\theoremstyle{thmstylethree}%

\newcommand*{\dif}{\mathop{}\!\mathrm{d}}

\begin{document}

\title[Article Title]{Evolutionary Spiking Neural Networks: A Survey}


\author[1,2]{\fnm{Shuaijie} \sur{Shen}}\email{12132355@mail.sustech.edu.cn}

\author[1,2]{\fnm{Rui} \sur{Zhang}}\email{12032455@mail.sustech.edu.cn}
\author[1,2]{\fnm{Chao} \sur{Wang}}\email{15662672289@163.com}
\author[1,2]{\fnm{Renzhuo} \sur{Huang}}\email{12332478@mail.sustech.edu.cn}
\author[3,2]{\fnm{Aiersi} \sur{Tuerhong}}\email{20211385@stu.cqu.edu.cn}
\author[2]{\fnm{Qinghai} \sur{Guo}}\email{guoqinghai@huawei.com}
\author[4]{\fnm{Zhichao} \sur{Lu}}\email{zhichao.lu@cityu.edu.hk}
\author*[1]{\fnm{Jianguo} \sur{Zhang}}\email{zhangjg@sustech.edu.cn}
\author*[2]{\fnm{Luziwei} \sur{Leng}}\email{lengluziwei@huawei.com}

\affil[1]{\orgdiv{Department of Computer Science and Engineering}, \orgname{Southern University of Science and Technology}, \orgaddress{\city{Shenzhen}, \country{China}}}

\affil[2]{\orgdiv{ACS Lab}, \orgname{Huawei Technologies}, \orgaddress{\city{Shenzhen}, \country{China}}}

\affil[3]{\orgdiv{College of Mathematics and Statistics}, \orgname{Chongqing University}, \orgaddress{\city{Chongqing}, \country{China}}}

\affil[4]{\orgdiv{Department of Computer Science}, \orgname{
City University of Hong Kong}, \orgaddress{\city{Hong Kong}, \country{China}}}


\abstract{Spiking neural networks (SNNs) are gaining increasing attention as potential computationally efficient alternatives to traditional artificial neural networks (ANNs). However, the unique information propagation mechanisms and the complexity of SNN neuron models pose challenges for adopting traditional methods developed for ANNs to SNNs. These challenges include both weight learning and architecture design. While surrogate gradient learning has shown some success in addressing the former challenge, the latter remains relatively unexplored. Recently, a novel paradigm utilizing evolutionary computation methods has emerged to tackle these challenges. This approach has resulted in the development of a variety of energy-efficient and high-performance SNNs across a wide range of machine learning benchmarks. In this paper, we present a survey of these works and initiate discussions on potential challenges ahead.}

\keywords{Spiking Neural Networks, Evolutionary Algorithm, Neural Architecture Search}



\maketitle

\section{Introduction}\label{sec:intro}
Spiking neural network (SNN) \cite{maass1997networks} has attracted increasing attention as a potential alternative to the traditional artificial neural network (ANN), with appealing attributes such as sparse computation and temporal dynamics. 
Recently, the problem associated with training deep SNNs has been arguably addressed by surrogate gradient (SG) methods where soft relaxed functions were used to approximate the original non-existing gradient of binary spiking activation \cite{bohte2000spikeprop,wu2018spatio,neftci2019surrogate}. 
Based on these methods, SNNs have achieved high-level performances in various deep learning tasks such as image classification on CIFAR \cite{cifar} and ImageNet \cite{deng2009imagenet} datasets \cite{wu2019direct,zheng2021going,zhou2022spikformer}. 
However, these approaches often adopted architectures optimized for ANNs and directly applied them to SNNs, which are likely to be sub-optimal for spike-based computation, resulting in unsatisfactory performance. 
In particular, the performance degradation was more apparent in tasks where variations in network architectures are required, such as dense image prediction \cite{zhu2022event,hagenaars2021self,kim2021beyond}.
To close this gap, a steady stream of works has been proposed, including improving SNN training and configuration of spike-based operators \cite{li2021differentiable,lian2023learnable,yao2023spike}. 
However, these methods largely relied on handcrafted design via trial and error. 
Alternatively, the recently emerged evolutionary approaches leveraged search algorithms to optimize network architectures and have been demonstrated to be efficient in constructing high-performance SNNs. When coupled with evolutionary algorithms (EAs), SNNs undergo automated optimization, adjusting hyperparameters, operations and architecture through evolutionary principles, while conventional SNNs are often handcrafted and only the connection weights are trained. Compared with the handcrafted SNNs, SNNs with EAs offer the advantages of enhanced flexibility, allowing them to meet different requirements in, e.g. computation cost, biological plausibility and hardware compatibility, etc.

This work provides a review of recent works of evolutionary SNNs. 
Previous review works on SNN have covered topics such as biological-inspired strategies \cite{nunes2022spiking}, learning rules \cite{nunes2022spiking,wang2022hierarchical}, energy efficiency \cite{malcolm2023comprehensive}, applications \cite{yamazaki2022spiking} and membrane computing theory \cite{paul2024survey}. Our work differs from these since it particularly discusses recent ideas of applying evolutionary algorithms to improve deep SNNs as an alternative to traditional handcrafted methods. We first introduce spiking neuron models which serve as fundamental building blocks in Section \ref{sec:neuron}. In Section \ref{sec:trainingSNN}, we briefly introduce the learning algorithms for SNNs, which are critical for realizing functional networks and highly involved in evolutionary methods in SNN. In Section \ref{sec:evolSNN}, we review recent works on evolutionary SNNs and provide a background introduction of evolutionary algorithms (EAs) in the beginning. Finally, we provide a comprehensive and quantitative comparison of recent works on benchmark image classification tasks and conclude with potential challenges of current evolutionary approaches for SNNs.

\section{Spiking Neuron Models}\label{sec:neuron}
In this section, we first introduce a general neuron model, Hodgkin–Huxley model \cite{hodgkin1952quantitative}, and then present some simplified spike neuron models commonly used in modern SNNs. Although a variety of spiking neuron models were proposed in prior works, only a few of them were adopted in modern deep SNNs due to computational efficiency and memory conservation.

\subsection{Hodgkin–Huxley Model}
Based on experiments conducted on the giant axon of a squid, the Hodgkin–Huxley (HH) model \cite{hodgkin1952quantitative} describes how action potentials in neurons are initiated and propagated. The model relates the variation of the membrane potential to the concentration of K$^+$ and Na$^+$:
\begin{equation}
C \frac{\dif u}{\dif t}  =  - \sum_k I_k (t) + I(t),
\end{equation}
where $C$ is membrane capacitance, $u$ is the membrane potential, $I$ is the current injected into cell, i.e. the synaptic input current, and $\sum_k I_k$ is the sum of the ionic currents which pass through the cell membrane. $\sum_k I_k$ is related to three types of channel which can be characterized by the conductance of the ion channels and the leakage channel, and is formulated by:
\begin{equation}
\sum_k I_k = g_{Na} m^3 h (u - E_{Na}) + g_{K} n^4 (u - E_{K}) + g_{L} (u - E_{L}),
\end{equation}
where $g_{Na}$, $g_{K}$, and $g_{L}$ are the maximum conductance of Na$^+$ channel, K$^+$ channel, and leakage channel, respectively. $E_{Na}$, $E_{K}$ and $E_{L}$ are the reversal potential. $m$, $h$ and $n$ are voltage-dependent "gating" variables to model the probability that a channel is open at a given moment in time.

Because of the demanding computation and complex simulation \cite{meunier2002playing} \cite{strassberg1993limitations}, HH model is not practical for large-scale tasks or deployment on ASIC, therefore, some simplified models are proposed to reduce the demand for computational resources.

\subsection{Leaky Integrate and Fire Model}
The leaky integrate-and-fire (LIF) model is an extremely simplified model that only remains the key points of voltage evolution and simplifies the detail of evolution. The voltage of leaky integrate-and-fire model evolves according to
\begin{equation}
\tau \frac{\dif}{\dif t} u = -(u - u_{rest}) + RI,
\end{equation}
where $\tau$ is the time constant, $u_{rest}$ is the membrane potential of the neuron at rest, $R$ is a voltage-dependent resistance and $I$ is the input current. When the voltage reaches a formal threshold, a spike is generated and then the voltage is instantly reset to a new value, as following describes
\begin{equation}
    u: \vartheta^{-} \to \vartheta, \, u \gets v_{r},
\end{equation}
where $\vartheta$ is the threshold which once reaches from below a spike is generated, and $v_{r}$ is the reset voltage. In modern SNNs, a discrete version of LIF neuron model is often adopted, described by:

\begin{equation}
    u_{t}' = u_{t - 1} - \frac{1}{\tau}(u_{t - 1} - u_{rest}) + x_{t+1}, 
\end{equation}
where $u_t'$ is the temporary voltage because the neuron will immediately determine whether to fire a spike based on the voltage.
\begin{equation}\label{eq:sgf}
    s_{t} = \mathbb I_{u_t' \ge \vartheta},
\end{equation}
where when $u_t' \ge \vartheta$, $s_t = 1$, otherwise, $s_t= 0$. And then the final value of voltage is determined based on $s_t$.
\begin{equation}
    u_{t} = (1 - s_t)u_t' + s_tu_r, 
\end{equation}
where the voltage is reset to a new value $u_r$ if the neuron generates a spike, which is also known as "hard reset", while deducing threshold from $u_t'$ is known as "soft reset", which is widely used in ANN-SNN conversion to pursue approximate equivalence.

LIF model, as its name suggests, its membrane voltage decays over time. But in some cases, a non-leaky model, also known as IF model will be used.

When coupled with the discrete LIF model, SNNs can be conceptualized as ANNs with multiple time-steps. In this framework, the activation of neuron in each time-step is influenced by the status of the activation in the previous time-step, reflecting the temporal dynamics of neural activity. In many cases, the final output of SNNs is computed as the mean of all outputs over all time-steps, and the loss function is computed using the final output. In certain cases, the loss function is also computed using the output from each time-step individually, and then aggregated to obtain an overall loss \cite{deng2022temporal}.

\section{Training of Spiking Neural Network}\label{sec:trainingSNN}

In this section, we briefly introduce the training methods of SNN which are highly involved in evolutionary SNN approaches. Towards training a brain-inspired neural network, learning methods inspired from biological nervous system are adopted, e.g., Spike-Timing-Dependent Plasticity (STDP), Anti-Hebbian STDP (aSTDP) and their Variants. For deep learning tasks, the training of SNNs can be mainly divided into two branches, i.e. ANN-to-SNN conversion and surrogate gradient training methods. The former is motivated by the success of ANNs, aiming to convert neurons with floating-value activation to spiking activation, based on weight transformation techniques. The later directly trains SNNs by approximating gradient-decent-based training methods with surrogate gradient functions.

\subsection{Spike-Timing-Dependent Plasticity}
Spike-Timing-Dependent Plasticity, a form of Hebbian learning \cite{hebb2005organization}, is an unsupervised learning method that adjusts the strength of connections (weight) between neurons based on the relative spike timing between them \cite{bi1998synaptic,song2000competitive}. More specifically, when the presynaptic neuron fires a spike before a postsynaptic neuron firing, the strength of the connection (synaptic weight) between them is increased, otherwise, it is decreased, which are known as long-term potential (LTP) \cite{bliss1973long} and depression (LTD) \cite{10.1371/journal.pone.0001377}, respectively. However, there are many neuronal systems not following Hebbian rule, which is known as anti-Hebbian STDP (aSTDP) that modifies the weight in a reverse way \cite{bell1997synaptic}. Besides, some STDP variants like Mirrored STDP (mSTDP) \cite{burbank2015mirrored}, Probabilistic STDP \cite{tavanaei2016acquisition} and Reward Modulated STDP (R-STDP) \cite{izhikevich2007solving} are introduced for specific purposes. Although these training methods are biologically inspired and easy to implement, they are yet to be demonstrated as competitive as back-propagation training \cite{rumelhart1986learning} for hard deep learning tasks \cite{journe2022hebbian}.

\subsection{ANN-SNN Conversion}

The difficulty of training SNNs and the high performances of ANNs in recent years have motivated researchers to convert the pre-trained ANNs to SNNs \cite{leng2014deep,cao2015spiking,leng2016spiking,diehl2015fast,rueckauer2017conversion,leng2018spiking}. Due to the intrinsic difference of information representation, SNNs obtained by early conversion-based methods often took dozens or even hundreds of steps to achieve competitive performance with the pre-trained ANNs \cite{sengupta2019going,Han_2020_CVPR,li2021free}. To this end, researchers recently proposed more efficient conversion methods including quantized activation functions by which the converted SNNs can achieve almost the same performance as ANNs in very few time steps \cite{bu2023optimal,pmlr-v202-jiang23a,ijcai2023p342}.

\subsection{Surrogate Gradient Learning Algorithm}

The success of ANNs relies heavily on back-propagation training \cite{rumelhart1986learning}, which also prompts researchers to using similar methods in SNNs. However, the intrinsic non-differentiability of the spike generation function (equation \ref{eq:sgf}) makes back-propagation impossible in training SNNs. To solve the problem, many works propose surrogate gradient (SG) functions to replace the non-differential term in back-propagation \cite{bohte2000spikeprop,wu2018spatio,mostafa2017supervised}. More specifically, the Dirac delta function, the real derivative of spike generation function, is replaced by a function of approximate shape, e.g. piece-wise quadratic function \cite{Esser_2016,wu2018spatio,bellec2018long} and sigmoid function \cite{bellec2018long,wozniak2020deep}. However, the SG method also introduces noise because of the mismatched gradient during the training process, making it difficult for SNNs to surpass ANNs. To this end, recent works proposed solutions such as adaptive SGs \cite{li2021differentiable,lian2023learnable,che2022differentiable}, membrane potential normalization  \cite{Guo_2023_ICCV,guo2022recdis} or improving spike-based operators \cite{yao2023spike}, etc. Although SNNs have high forward inference speed and theoretically low energy consumption due to sparse computation, they require a relative high amount of memory and computation during training compared to ANNs because of the extra temporal dimension. To alleviate these problems, recent works reduced the amount of calculation in back-propagation by taking advantage of the sparsity of SNNs \cite{perez2021sparse,xiao2022online,Meng_2023_ICCV}.

\section{Evolutionary Spiking Neural Networks}\label{sec:evolSNN}
In this section, we first make a briefly introduction on EAs as background knowledge and focus on methods that have been applied to SNNs. We then present and analyse recent works of evolutionary SNNs. In the end, we summarize and provide a comprehensive comparison of recent deep evolutionary SNNs on benchmark image classification tasks.
\subsection{Evolutionary Algorithms}
Drawing inspiration from the biological process of evolution, EAs are effectively utilized in fields such as optimization, learning, and design. These algorithms operate on the principle of a population-based generate-and-test method. The generation phase involves the mutation and/or recombination of individuals within a population, emulating natural evolutionary processes. Subsequently, in the testing phase, the algorithm selects the next generation by choosing from both the parents and their offspring, based on their performance or fitness. In this continuous process of generating and testing, the evolutionary algorithm iteratively refines the population until a satisfying solution is identified, adhering to predefined halting criteria.

A simple evolutionary algorithm is described in Algorithm \ref{algo:Evolutionary Algorithm}.
\begin{algorithm}
    \caption{Evolutionary Algorithm}
    \label{algo:Evolutionary Algorithm}
    \begin{algorithmic}[1]
    \State Initialize the population $P=\{x^1, x^2, ... ,x^n\}$ and calculate their fitness;
    \While{The termination criteria is not met}
      \State Select parents $P_{parents}$ from the population $P$;
      \State Reproduce children $P_{child}$ from $P_{parents}$ by mutation and crossover;
      \State Evaluate fitness of $P_{child}$;
      \State Let individuals in $P$ and $P_{child}$ compete based on fitness, then select survived individuals as new population $P'$, and assign $P'$ to $P$;
    \EndWhile
    \State \Return{final population $P$}
    \end{algorithmic}
	
\end{algorithm}

The population $P$, contains a set of individuals $x^1, x^2, ... ,x^n$ with $n$ denoting the total number of individuals in the population. Each individual is represented by a string termed \emph{chromosome} or \emph{genotype}.
$P_{parents}$ represents the chosen parents from the current population $P$. Usually, individuals with higher fitness are more likely to be selected.
$P_{child}$ are generated from $P_{parents}$ using mutation and crossover operators, which are key methods in evolutionary algorithms to create new individuals. The fitness is a measure of how well an individual ($P_{child}$) adapts to the environment. This loop continues until a termination criterion is met, e.g. reaching a maximum number of iterations or finding a sufficiently good solution.

Essentially, EAs are not just one algorithm, but a whole family of different algorithms, each with its own history, methods and strategies.  
Examples include Genetic Algorithms (GAs)\cite{holland1992genetic} that use natural selection concepts, Evolutionary Programming (EP)\cite{fogel1962autonomous} which focuses on evolving behaviors, Evolution Strategies (ES)\cite{rechenberg1965cybernetic,schwefel1965kybernetische} for optimizing real values, Genetic Programming (GP)\cite{de1990genetic,koza1990genetic} that evolves computer programs, Differential Evolution (DE)\cite{storn1996usage} known for its straightforward optimization approach, and Particle Swarm Optimization (PSO)\cite{kennedy1995particle} inspired by the social behavior of birds and fish. Each of these represents a unique approach within the broad scope of EAs, leading to their widespread application in various specialized fields. 

\subsection{Evolutionary SNNs}

The inherent complexity of SNNs naturally provides diverse optimization objectives. Early works have applied various EAs, such as GA, DE, PSO and harmony search algorithms on SNNs, optimizing hyperparameters including network topology, spiking neurons, synaptic weights and delay \cite{pavlidis2005spiking,batllori2011evolving,vazquez2011training,saleh2014novel,schaffer2015evolving,yusuf2017evolving}, etc. However, limited by the hardware computing power, only shallow SNNs and small scale datasets were adopted in experiments, with tasks ranging from classification to robotic control. Boosted by the hardware and software advancements of deep learning, more recent evolutionary SNN works have extended to deeper networks and competed with both handcrafted SNNs and benchmark ANNs. We categorize these works into neural architecture search (NAS) and non-NAS methods.

\subsubsection{NAS methods}

AutoSNN \cite{na2022autosnn} proposed a spike-aware neural architecture search (NAS) framework and applied to image classification on CIFAR and Tiny-ImageNet datasets. The work defines a two-level search space on backbone architecture and candidate block sets. It further analyzes the effects of the architecture components on the accuracy and number of spikes, which suggests using max pooling over global average pooling as down-sampling layers in SNNs. To reduce search cost induced by training and evaluating candidate architectures, the work adopts a one-shot weight-sharing approach \cite{pham2018efficient,cai2019once} using evolutionary algorithm. The work also proposes a spike-aware fitness function to penalize architectures with more spikes. Based on the fitness function, an evolutionary algorithm is performed on the trained supernet. New populations are generated using mutation and crossover by the parent population. Finally, the best architecture with the highest fitness value is selected, achieving an energy-efficient SNN with low spike rate and competitive performance. However, the algorithm is relatively complex and takes a long time to train the supernet. Additionally, the single-path sampling makes the training of spiking blocks unstable. 

Concurrent to AutoSNN, \cite{kim2022neural} proposed SNASNet to find optimal architectures using NAS without training the SNN, which typically requires significantly longer training time compared to ANNs. Motivated by the observation that network architecture with a high representation power at initialization is likely to achieve higher
post-training accuracy \cite{mellor2021neural,chen2021neural}, SNASNet finds an SNN-friendly architecture by selecting the architecture that can represent diverse spike activation patterns across different data samples without training. To this end, the work proposes Sparsity-Aware Hamming Distance (SAHD) for addressing sparsity variation of LIF neurons. In addition, to explore temporal information represented by spikes, the work searches for both the forward as well as backward connections between layers, and achieves higher performance with backward connections. 

Following the weight sharing approach, SpikeDHS \cite{che2022differentiable} adopted a gradient-based differentiable architecture search (DARTS) \cite{liu2018darts,Liu_2019_CVPR} method to search SNNs. The supernet is trained end-to-end in a continuous relaxation search space where both the cell and layer structure are optimized alternately with the model weights. In addition, the work ensures multiplication-free inference (MFI) \cite{li2023efficient} which is essential for SNNs to achieve low power computing, by defining the cell operation with spike-based computation. To solve the noise problem induced by fixed SGs, in the retraining phase, a differentiable surrogate gradient search (DGS) method is proposed which evolves the SG function by adjusting its shape periodically. The searched SNNs are applied to static image classification on CIFAR and ImageNet datasets, as well as challenging dynamic event-based deep stereo tasks on the MVSEC dataset \cite{zhu2018multivehicle}, both achieving high performances meanwhile with much lower energy consumption than ANNs. The searched SNN encoder topology is also applied to event-based object detection \cite{zhang2023automotive}, demonstrating the versatility and efficiency of the method. A limitation of the work is that since the searched candidates are mixed by a weighted average, when difference between the coefficients is not significant, the decoded architecture could be suboptimal, which is observed in classification tasks. However, for tasks which require critical structure variation, such as dense prediction, this limitation is not notable, which is also demonstrated in extended works based on SpikeDHS \cite{zhang2023accurate}. 

Compared to previous NAS works on SNNs, \cite{shen2023brain} considered more biological principles in the search space and proposed the brain-inspired neural circuits evolution  (NeuEvo) strategy, which involves feedback connection, excitatory and inhibitory neurons \cite{suzuki2012microcircuits} and local learning. The work adopts an alternative bi-level optimization approach on synaptic weights and structure coefficients similar to previous works \cite{liu2018darts,che2022differentiable}. It further exploits the local spiking behavior of neurons to evolve neural circuits with STDP and updates the synaptic weights in combination with gradient-based global error signals. The evolved SNNs are applied to static and event-based image classification, and reinforcement learning tasks, achieving competitive performances to ANNs meanwhile with highly sparse network activation.

For the purpose of a better accuracy-computational cost trade-off, \cite{yan2024efficient} adopted a single-path NAS \cite{stamoulis2019single} approach, encoding all candidate architectures in a branchless spiking supernet, which requires much less computation and thus significantly reduces the search time. Rather than training independent kernels with different kernel sizes, the framework trains a branchless superkernel inspired by \cite{stamoulis2019single} whose actual receptive field is controlled by trainable parameters. In addition, synaptic operation (SynOps)-aware optimization is proposed to find a computationally efficient subspace of the supernet. 

\subsubsection{Non-NAS methods}

To reduce the computation cost of the search process, \cite{yan2024sampling} proposed spatio-temporal topology sampling (STTS) algorithm for SNN which samples both complex spatial topology and temporal topology by incorporating the synaptic delay. The work adopts a multi-stage downsampling structure and spiking convolution node design similar to SpikeDHS \cite{che2022differentiable}. Differently, light-weight depthwise separable convolution \cite{chollet2017xception} is used to reduce model parameters. Furthermore, the topology of the network is not optimized by gradient-based learning rules, but sampled by random graph models for spatial topology and from a pre-defined distribution for temporal topology. The sampled architecture is directly used as the final architecture without training and evaluation, thus drastically reducing the computation cost compared with other NAS methods. However, as mentioned in the work, there is still a lack of theoretical guarantee of the random sampling method, although empirical results show that the variance between different sampled architectures is low. 

\cite{wang2023evolving} proposed the evolving connectivity (EC) framework to train recurrent SNNs (RSNNs), which employs natural evolution strategies (NES) \cite{JMLR:v15:wierstra14a} for optimizing parameterized connection probability distributions to replace the weight-tuning process. Inspired from weight agnostic network \cite{gaier2019weight}, the EC framework circumvents the need for gradients and features hardware-friendly characteristics, including sparse binary connections and accelerated training. The evolved SNNs are evaluated on a series of robotic locomotion tasks including a complex 17-DoF humanoid task, where it achieves comparable performance with ANNs and outperformed gradient-trained RSNNs.

\subsubsection{Discussion}

Finally, we provide results of above works on benchmark image classification tasks including CIFAR10, CIFAR100, CIFAR10-DVS, Tiny-ImageNet and ImageNet datasets in Table \ref{tab:performance_comparison}. For a comprehensive comparison, we adopt evaluation metrics including number of model parameters (NoP), simulation time steps (T), accuracy, number of spikes (NoS), theoretical power consumption and synaptic operations (SynOps). Note that the theoretical power consumption is calculated by operation numbers of the SNN during the inference of one image, we refer to \cite{yan2024sampling} for a nice elaboration of the relation between SynOps and theoretical power consumption. We also include recent state-of-the-art works of directly handcrafted SNNs as a comparison with evolutionary approaches. 

On CIFAR dataset series, evolutionary SNNs are highly competitive to directly handcrafted SNNs. For convolution networks, the precision of SpikeDHS(n4) \cite{che2022differentiable} surpasses the handcrafted Dspike \cite{li2021differentiable} on CIFAR10 and CIFAR100 under similar network size. With DGS method, SpikeDHS$^D$(n3c5) is even competitive to the Spike-driven Transformer \cite{yao2024spike} which adopts a transformer architecture. In terms of accuracy, NeuEvo \cite{shen2023brain} currently achieves the highest values among evolutionary SNNs on CIFAR10, CIFAR100 and CIFAR10-DVS datasets. On the large scale ImageNet dataset, among convolution networks, STTS \cite{yan2024sampling} demonstrates the highest accuracy within evolutionary SNNs meanwhile significantly outperforming the handcrafted Dspike \cite{li2021differentiable} under similar network size. However, the very recent work of Meta-SpikeFormer \cite{yao2023spike} using optimized spike-based attention operations achieves significantly higher accuracy than STTS meanwhile with much smaller network size, though with higher theoretical power consumption. This indicates the limitation of evolutionary SNN approaches based on pure convolution architectures.

\begin{table}[!ht]
    \centering
    \caption{Performance comparison of evolutionary SNNs and state-of-the-art handcrafted SNNs on image classification. All values are taken from literatures. Values by estimation are denoted by *.  Values not available are denoted by -. Handcrafted SNNs are denoted by bold prefix H.}
    \label{tab:performance_comparison}
    \begin{tabular}{@{}llcclccc@{}} 
        \toprule
        Dataset & Model & \begin{tabular}[c]{@{}c@{}}NoP\end{tabular} & \begin{tabular}[c]{@{}c@{}}T\end{tabular} & \begin{tabular}[c]{@{}c@{}}Accuracy\\ (\%)\end{tabular} & NoS &  \begin{tabular}[c]{@{}c@{}}Power\\ (mJ)\end{tabular} & \begin{tabular}[c]{@{}c@{}}SynOps
    \end{tabular} \\
    \midrule
        \multirow{10}{*}{CIFAR10} 
        & AutoSNN (C=128) \cite{na2022autosnn} & 21M & 8 & 93.15 & 310K & 6.3 & 732M\\      
        & SNASNet-BW \cite{kim2022neural} & 49.93M & 8 & 94.12$^{\pm 0.25}$ & - & - & -\\   
        & SpikeDHS-CLA(n4) \cite{che2022differentiable} & 12M & 6 & 94.34$^{\pm 0.06}$ & 788K & 5.5 & 923M\\
        & SpikeDHS-CLA(n3) \cite{che2022differentiable} & 14M & 6 & 95.35$^{\pm 0.05}$ & 752K & - & -\\       
        & SpikeDHS-CLA$^{D}$(n3c5) \cite{che2022differentiable} & 14M & 6 & 95.50$^{\pm 0.03}$ & 720K & - & - \\
        & STTS(TANet-Tiny) \cite{yan2024sampling} & 7M & 4 & 95.10$^{\pm 0.09}$ & - & 1.2 & -\\
        & ESNN ($\lambda$=0) \cite{yan2024efficient} & 23.47M & 3 & 94.64 & -& 1.2 & 412M \\
        & ESNN ($\lambda$=0.1) \cite{yan2024efficient} & 16.18M & 3 & 94.27 & -& 0.8 & 270M \\
        & NeuEvo \cite{shen2023brain} & - & - & 96.43$^{\pm 0.26}$ & 500K$^*$ & - & -\\
        & \textbf{(H)} Dspike \cite{li2021differentiable} & 11M & 6 & 94.25 & - & - & -\\
        & \textbf{(H)} Spike-driven Transformer \cite{yao2024spike} & - & 4 & 95.6 & - & - & -\\
   \midrule
        \multirow{9}{*}{CIFAR100} 
        & AutoSNN(C=64) \cite{na2022autosnn} & 5M & 8 & 69.16 & 326K & - & 785M\\
        & SNASNet-BW \cite{kim2022neural} & 20.71M & 5 & 73.04$^{\pm 0.36}$ & - & -& 1293M \\   
        & SpikeDHS-CLA(n4) \cite{che2022differentiable} & 12M & 6 & 75.70$^{\pm 0.14}$ & 962K & - & 1056M \\
        & SpikeDHS-CLA(n3) \cite{che2022differentiable} & 14M & 6 & 76.15$^{\pm 0.20}$ & -& - & - \\
        & SpikeDHS-CLA$^D$(n3s1) \cite{che2022differentiable} & 14M & 6 & 76.25$^{\pm 0.10}$ & - & -& - \\
        & STTS(TANet-Tiny) \cite{yan2024sampling} & 7M & 4 & 76.33$^{\pm 0.32}$  & - & - & -\\
        & ESNN ($\lambda$=0) \cite{yan2024efficient} & 27.55M & 3 & 74.78 & - & -& 517M \\
        & ESNN ($\lambda$=0.1) \cite{yan2024efficient} & 17.32M & 3 & 73.21 & - & -& 348M \\
        & NeuEvo \cite{shen2023brain} & - & - & 77.72$^{\pm 0.32}$ & - & - & -\\
        & \textbf{(H)} Dspike \cite{li2021differentiable} & 11M & 6 & 74.24$^{\pm 0.10}$ & - & - & -\\
        & \textbf{(H)} Spike-driven Transformer \cite{yao2024spike} & - & 4 & 78.4 & - & - & -\\
    \midrule
		\multirow{4}{*}{CIFAR10-DVS} 
        & AutoSNN(C=16) \cite{na2022autosnn} & 0.42M & 20 & 72.50 & 1269K & -& 1221M \\
		& ESNN ($\lambda$=0) \cite{yan2024efficient} & 3.53M & 10 & 78.40 & - & -& 1998M \\
        & ESNN ($\lambda$=0.1) \cite{yan2024efficient} & 2.52M & 10 & 75.80 & - & -& 1215M \\
        & NeuEvo \cite{shen2023brain} & - & - & 84.17$^{\pm 0.23}$ & - & - & -\\
        & \textbf{(H)} Dspike \cite{li2021differentiable} & 11M & 10 & 75.4$^{\pm 0.10}$ & - & - & -\\
        & \textbf{(H)} Spike-driven Transformer \cite{yao2024spike} & - & 16 & 80.0 & - & - & -\\
    \midrule
        \multirow{3}{*}{Tiny-ImageNet} 
        & AutoSNN(C=64) \cite{na2022autosnn} & 14.83M & 8 & 46.79 & 680K & - & 903M\\
		& SNASNet-BW \cite{kim2022neural} & 74.62M & 5 & 54.60$^{\pm  0.48}$ & - & - & 1454M\\   
        & ESNN ($\lambda$=0) \cite{yan2024efficient} & 6.91M & 3 & 58.59 & - & -& 1061M \\
		& ESNN ($\lambda$=0.1) \cite{yan2024efficient} & 5.14M & 3 & 58.10 & - & -& 666M \\
    \midrule
        \multirow{4}{*}{ImageNet} 
        & SpikeDHS \cite{che2022differentiable} & 58M & 6 & 67.96 & - & - & -\\
        & SpikeDHS$^{D}$ \cite{che2022differentiable} & 58M & 6 & 68.64 & - & -& - \\
        & STTS(TANet-Regular) \cite{yan2024sampling} & 25M & 4 & 70.79$^{\pm 0.43}$ & - & 8.2 & - \\
        & NeuEvo \cite{shen2023brain} & - & - & 68.74 & - & -& - \\            
        & \textbf{(H)} Dspike \cite{li2021differentiable} & 22M & 6 & 68.19 & - & - & -\\
        & \textbf{(H)} Meta-SpikeFormer \cite{yao2023spike} & 15.1M & 4 & 74.1 & - & 16.7 & - \\             
        & \textbf{(H)} Meta-SpikeFormer \cite{yao2023spike} & 55.4M & 4 & 79.7 & - & 52.4 & - \\      
    \bottomrule
    \end{tabular}
\end{table}

 \section{Summary}\label{sec:sum}
This paper provides a survey of emerging works of evolutionary SNNs, with a focus on recent approaches applied for deep SNNs. In general, current evolutionary methods enable SNNs to achieve high accuracy in tasks including static and event-based image classification, event-based vision tasks and reinforcement learning tasks in robotics, meanwhile with low energy consumption, evaluated by number of spikes, theoretical powers or SynOps. Despite recent achievements, limitations exit for current evolutionary approaches for SNNs. We conclude these challenges as following:
\begin{enumerate}
    \item Evolution cost: Several evolutionary SNN works \cite{na2022autosnn,che2022differentiable,shen2023brain} integrate gradient-based training into the architectural optimization process. Considering the training of SNN typically requires significantly longer time than ANN, this strategy could lead to huge computation cost when scaling up. This motivates efficient SNN evolutionary methods towards better accuracy-computational cost trade-off. Recent works have improved it using methods including initialization optimization, random sampling and branchless method \cite{kim2022neural,yan2024sampling,yan2024efficient}.
    \item Operation space: Current evolutionary SNN works mainly define their operation space on convolution, which largely limit their performances when competing with SNNs with attention mechanisms. This motivates a more flexible and diverse operation search space design compatible with spike-based operation, such as MLP \cite{li2023efficient} and attention \cite{che2023auto}.
    \item Co-evolution: Current works mainly focus on architecture optimization. Biological nervous systems utilize highly diverse neurons, network topology, learning rules and targets for optimization, to form intelligence. The integrated evolution of multiple elements such as architecture, neuron dynamics and learning algorithms could potentially lead to more efficient and powerful bio-inspired SNNs.
\end{enumerate} 

Alternative to improving the performance of SNNs on benchmark datasets, a more fundamental impact of EAs to SNNs is that their bio-mimetic nature offers a bottom-up path for the automated design of SNNs. The success of deep learning has motivated ANN-like SNNs with simplified LIF neuron model. However, biological neurons and neural circuits are featured with abundant dynamics and diverse interaction rules of local components, which resulting in a huge degree of freedom hard for handcrafted methods. Embedded with these principles, network combined with EAs could evolve through competition and mutation, towards intelligent systems with biological plausibility and answer underlying questions, such as why spikes?
\backmatter

\section*{Declarations}

\begin{itemize}
\item Funding: This work was supported by the National Key Research and Development Program of China (2021YFF1200800) and the Science and Technology Innovation (STI) 2030-Major Project (Brain Science and Brain-Like Intelligence Technology) under Grant 2022ZD0208700.
\item Conflict of interest/Competing interests: The authors declare that they have no conflict of interest.
\item Ethics approval and consent to participate: Not applicable
\item Consent for publication: All authors consent for publication.
\item Data availability: Not applicable
\item Materials availability: Not applicable
\item Code availability: Not applicable
\item Author contribution: Luziwei Leng, Shuaijie Shen and Zhichao Lu concieved the idea of the paper. Shuaijie Shen, Rui Zhang, Chao Wang, Renzhuo Huang, Aiersi Tuerhong, Qinghai Guo, Zhichao Lu and Luziwei Leng participated in the writing of the paper. Luziwei Leng and Jiangguo Zhang supervised the project. All authors reviewed the manuscript.
\end{itemize}




\bibliography{sn-bibliography}

\end{document}